\documentclass[sigconf]{acmart}

\usepackage{nopageno}
\usepackage{booktabs} 
\usepackage{float}
\usepackage{booktabs}
\usepackage{amsmath}
\usepackage{amsthm}
\usepackage{enumitem}
\usepackage{color}
\usepackage[utf8]{inputenc} 
\usepackage[T1]{fontenc}    
\usepackage{hyperref}       
\usepackage{url}            
\usepackage{booktabs}       
\usepackage{amsfonts}       
\usepackage{nicefrac}       
\usepackage{microtype}      
\usepackage{url}
\usepackage{graphicx}
\usepackage{adjustbox}
\usepackage{array,multirow}

\usepackage{caption}
\usepackage{subcaption}
\usepackage[utf8]{inputenc} 
\usepackage[T1]{fontenc}    
\usepackage{hyperref}       
\usepackage{url}            
\usepackage{booktabs}       
\usepackage{amsfonts}       
\usepackage{nicefrac}       
\usepackage{microtype}      
\usepackage{tablefootnote}
\usepackage[flushleft]{threeparttable}

\setcopyright{acmlicensed}

\acmConference[KDD'18]{Knowledge Discovery and Data Mining}{August 2018}{
  Longdon, United Kingdom}
\acmYear{2018}
\newcommand{\bz}{\mathbf{z}}
\newcommand{\bw}{\mathbf{w}}
\newcommand{\bx}{\mathbf{x}}

\begin{document}
	
\copyrightyear{2018} 
\acmYear{2018} 
\setcopyright{acmlicensed}
\acmConference[KDD '18]{The 24th ACM SIGKDD International Conference on Knowledge Discovery \& Data Mining}{August 19--23, 2018}{London, United Kingdom}
\acmBooktitle{KDD '18: The 24th ACM SIGKDD International Conference on Knowledge Discovery \& Data Mining, August 19--23, 2018, London, United Kingdom}
\acmPrice{15.00}
\acmDOI{10.1145/3219819.3220102}
\acmISBN{978-1-4503-5552-0/18/08}

\setlength{\abovedisplayskip}{3pt}
\setlength{\belowdisplayskip}{3pt}
\title{Deep Multi-Output Forecasting}
\subtitle{Learning to Accurately Predict Blood Glucose Trajectories}


	\author{Ian Fox$^1$, Lynn Ang$^2$, Mamta Jaiswal$^2$, Rodica Pop-Busui$^2$, Jenna Wiens$^1$}
	\affiliation{
		\institution{$^1$CSE, University of Michigan, $^2$Internal Medicine, University of Michigan}
	}

	\renewcommand{\shortauthors}{I. Fox et al.}


\begin{abstract}
\thispagestyle{empty}
  In many forecasting applications, it is valuable to predict not only the value of a signal at a certain time point in the future, but also the values leading up to that point. This is especially true in clinical applications, where the future state of the patient can be less important than the patient's overall trajectory. This requires multi-step forecasting, a forecasting variant where one aims to predict multiple values in the future simultaneously. Standard methods to accomplish this can propagate error from prediction to prediction, reducing quality over the long term. In light of these challenges, we propose multi-output deep architectures for multi-step forecasting in which we explicitly model the distribution of future values of the signal over a prediction horizon. We apply these techniques to the challenging and clinically relevant task of blood glucose forecasting. Through a series of experiments on a real-world dataset consisting of 550K blood glucose measurements, we demonstrate the effectiveness of our proposed approaches in capturing the underlying signal dynamics. Compared to existing shallow and deep methods, we find that our proposed approaches improve performance individually and capture complementary information, leading to a large improvement over the baseline when combined (4.87 \textit{vs.} 5.31 absolute percentage error (APE)). Overall, the results suggest the efficacy of our proposed approach in predicting blood glucose level and multi-step forecasting more generally.

  \begin{small}
  \vspace{3pt}
  \noindent\textbf{ACM Reference Format:}

  \noindent Ian Fox, Lynn Ang, Mamta Jaiswal, Rodica Pop-Busui, Jenna Wiens. 2018. Deep Multi-Output Forecasting: Learning to Accurately Predict Blood Glucose Trajectories. In \textit{KDD ’18: The 24th ACM SIGKDD International Conference on Knowledge Discovery \& Data Mining}, August 19–23, 2018, London, UK. ACM, New York, NY, USA, 9 pages. \url{https://doi.org/10.1145/3219819.3220102}
  \end{small}

\end{abstract}


\settopmatter{printfolios=true}

\maketitle

\section{Introduction}
In a typical signal forecasting problem, one aims to estimate the future value of the signal using past values. For example, one may aim to predict a blood glucose measurement occurring 30 minutes in the future, given past blood glucose measurements. This single-step setting generalizes to the multi-step setting, in which one aims to predict multiple values within a time horizon. This multi-step setting is inherently more difficult, since it requires modeling the joint probability of future measurements. While more challenging, if successful this joint modeling of observations within a sequence can improve overall performance. For example, while the word `the' occurs often in English, the phrase `the the the' does not. 

Recursive approaches, in which a single-step forecaster predicts several values by using the current prediction to make the next prediction, are commonly used in multi-step forecasting \cite{taieb_review_2012}. However, such approaches often suffer from poor long term performance, since any error introduced will enter a positive feedback loop. Alternatively, multi-output forecasting aims to estimate multiple values at once. While no longer susceptible to the feedback issue, multi-output forecasting may not adequately capture dependencies among predictions. 

We propose two complementary solutions to these issues. The first is a multi-output recurrent neural network where explicit temporal dependencies between outputs capture the relationship between the predictions. The second is a novel architecture that directly models the underlying generating function of the signal by learning a polynomial approximation for the outputs. The problem of error accumulation during sequence prediction has been previously studied in NLP \cite{bengio_scheduled_2015, lamb_professor_2016}. We distinguish ourselves from this past work by focusing on new models that alleviate this problem, as opposed to new training schemes.

We apply the proposed approaches to a challenging real-world forecasting problem (described below). Our main contributions can be summarized as followed:

\begin{itemize}
	\item We propose two novel and complementary deep multi-output forecasting architectures: an autoregressive multi-output forecaster and a polynomial ``function forecasting'' system.
	\item We improve over existing approaches by leveraging the proposed forecasting architectures on a large-scale real-world multi-step forecasting problem.
\end{itemize}

In additional analyses, we demonstrate that predicting multiple values can provide extra supervision, improving single output forecasting performance.

This work focuses on forecasting blood glucose values. Forecasting blood glucose is relevant to individuals with type 1 diabetes. In the United States alone, there are over 1 million type 1 diabetics \cite{tao_estimating_2010}. Tight glucose control, which can reduce the risk of complications in diabetes \cite{noauthor_effect_1993, writing_team_for_the_diabetes_control_and_complications_trial/epidemiology_of_diabetes_interventions_and_complications_research_group_sustained_2003, noauthor_intensive_2011, nathan_diabetes_2013}, can be challenging to maintain. There is a tremendous decision burden placed on diabetic patients who are constantly faced with decisions pertaining to food intake, activities, and insulin administration. Thus there is a  continuous interest in the field to develop sensitive technologies and algorithms to close the loop in insulin delivery and continouos glucose monitoring \cite{cobelli_artificial_2011}. Better predictive algorithms are critical in the development of such technologies \cite{man_uva/padova_2014}. This work could also be useful for type 2 diabetics with poor blood glucose control.

Learning the dynamics of the glucoregulatory system is difficult because the long term system dynamics are highly nonlinear \cite{hovorka_nonlinear_2004}. We tackle this challenge using data from over 550K blood glucose measurements. Our proposed approaches, when used together, achieve better results in terms of Absolute Percent Error (APE) than existing shallow or deep forecasting approaches both on average (4.87 \textit{vs.} 5.31) and particularly in periods of extreme fluctuation (12.05 \textit{vs.} 13.34). 

The remainder of the paper is organized as follows. In the next section, we introduce notation and formally define our problem. We then present our proposed forecasting architectures and methods. After, we present a series of experiments on the real-world dataset, and discuss the results. 

\section{Problem Setup and Background}
In signal forecasting, one aims to estimate the next value in a signal $x_{t+1}$ given past values $\mathbf{x}_{0:t}$, $\mathbf{x}$ represents the signal of interest, and $t$ the current time step. Here, we focus on the univariate setting, \textit{i.e.}, $x_{0},\dots,x_{t}\in\mathbb{R}$, though our approaches generalize to the multidimensional setting. The most common approaches to signal forecasting focus on learning a model for $p(x_{t+1}|\mathbf{x}_{0:t})$ \cite{oord_wavenet:_2016,taieb_review_2012}. Sometimes a prediction offset $d$ is added to learn the model for $p(x_{t+d}|\mathbf{x}_{0:t})$. Given the recent successes of deep architectures for this problem in general \cite{gensler_deep_2016,oord_wavenet:_2016,wu_probabilistic_2016}, and specifically in the domain of glucose forecasting \cite{mirshekarian_using_nodate}, we focus on building upon deep learning methods for signal forecasting.

\begin{figure}
	\centering
	\includegraphics[width=0.9\linewidth]{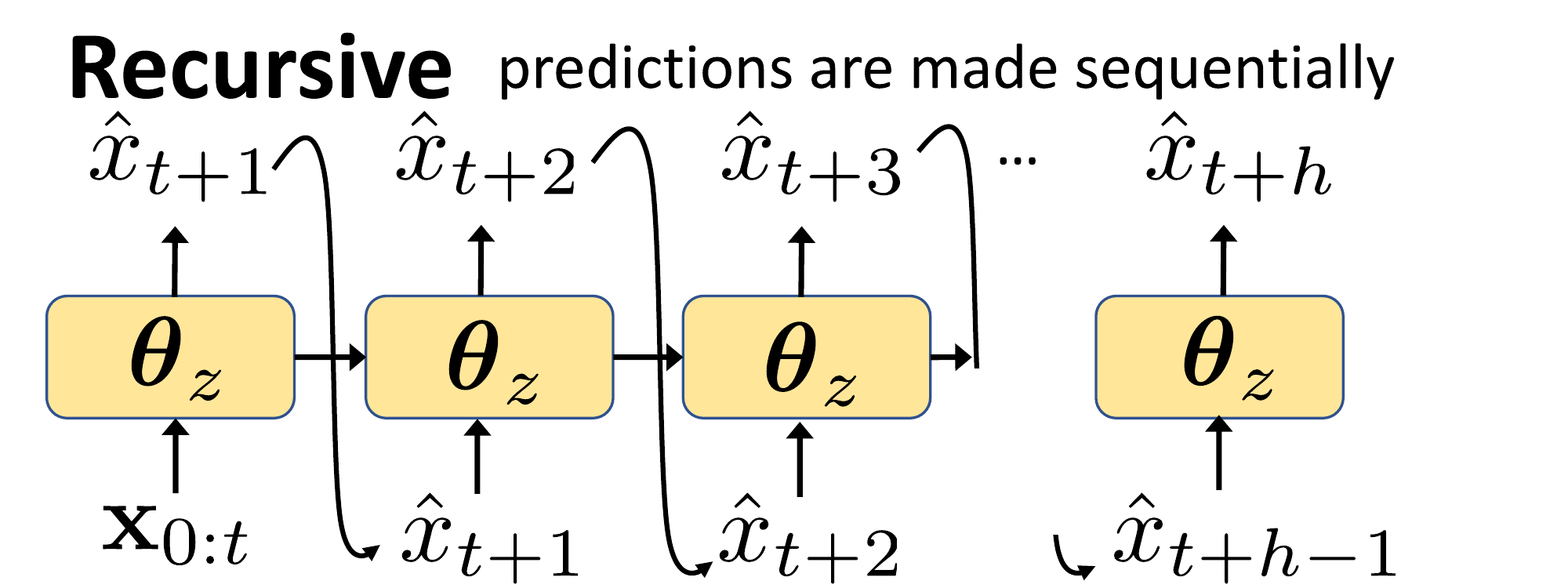}
	\caption{An example of multi-step recursive forecasting. Predictions at one time step are fed back into the network as input. This allows for single-step methods to produce multi-step forecasts.}
	\label{fig:new_rec}
\end{figure}

A model that accurately predicts $x_{t+1}$ can be used for either single- or multi-step forecasting. Applied recursively, single-step models enable multi-step forecasting, \textit{i.e.}, predicting multiple values over a time horizon of length $h$ (see \textbf{Figure \ref{fig:new_rec}}). Of particular note are deep conditional generative models, which model joint distributions by sequentially estimating terms in the conditional factorization of the distribution with deep neural nets \cite{oord_wavenet:_2016}. This style of forecasting, where $p(\mathbf{x}_{t+1:t+h}|\mathbf{x}_{0:t})$ is estimated by the factorization:
$p(x_{t+1}|\mathbf{x}_{0:t})p(x_{t+2}|\mathbf{x}_{0:t},\hat{x}_{t+1}) \dots p(x_{t+h}|\mathbf{x}_{0:t}, \hat{\mathbf{x}}_{t+1:t+h-1})$
is called \textbf{recursive forecasting}, and is the primary form of multi-step forecasting \cite{taieb_review_2012}. Recursive models have the advantage of modeling the joint probability of the signal within the prediction window. However, the re-use of predictions creates a feedback loop, amplifying potential errors and leading to lower quality predictions as the time horizon increases. 

In contrast, \textbf{multi-output} forecasting aims to estimate \\
$p(\mathbf{x}_{t+1:t+h}|\mathbf{x}_{0:t})$ in one step \cite{taieb_review_2012}. Multi-output approaches sidestep the issue of error feedback by jointly estimating over the prediction window, and will be the main focus of this paper.

\section{Methods}
\begin{figure}
\centering
\includegraphics[width=0.9\linewidth]{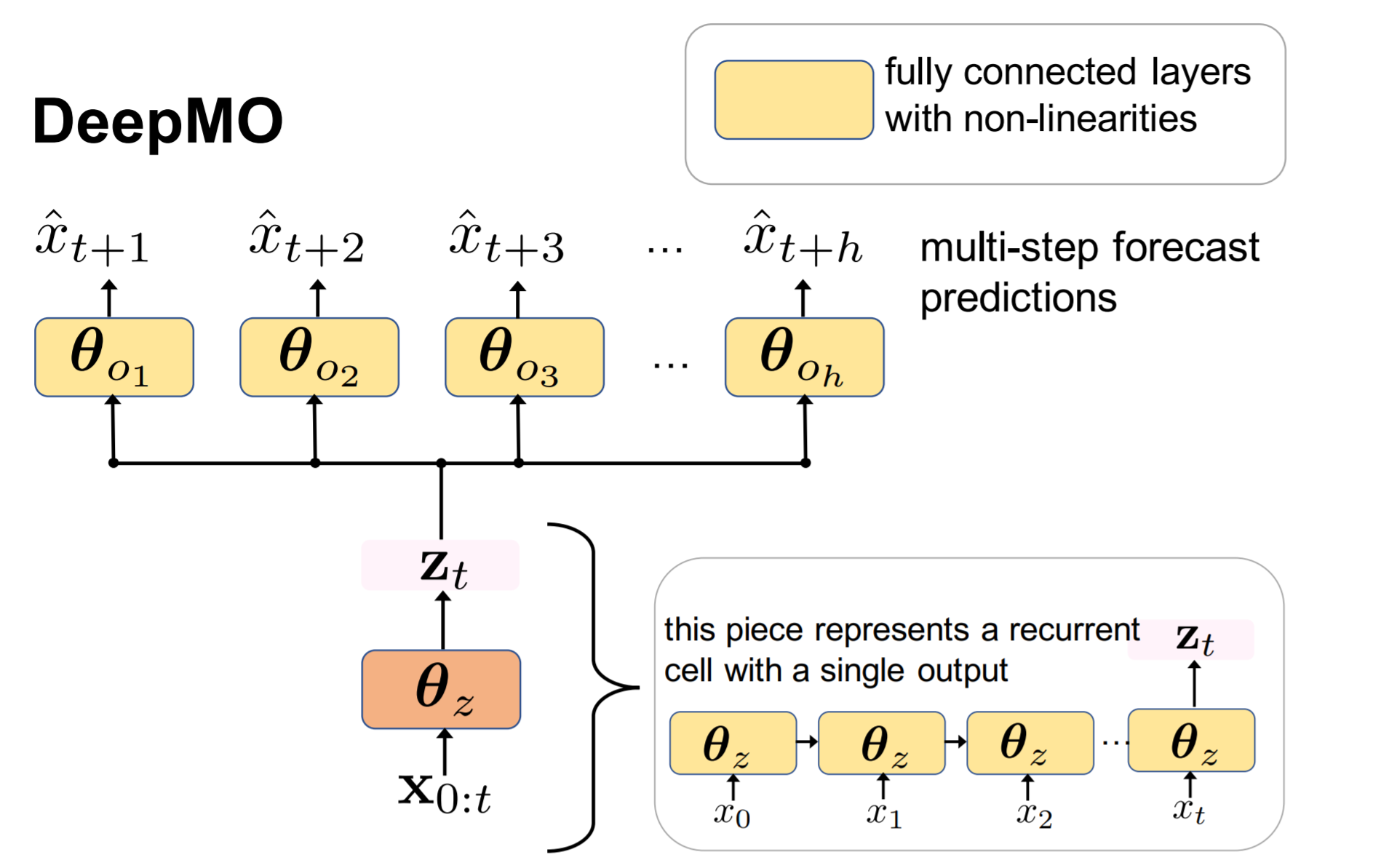}
\caption{Forecasting with DeepMO involves transforming the input to a shared representation and then learning separate output networks for each time point in the prediction horizon $h$.}
\label{fig:DeepMO}
\end{figure}

\begin{figure*}
\centering
\begin{subfigure}{0.49\textwidth}
  \centering
\includegraphics[width=1\linewidth]{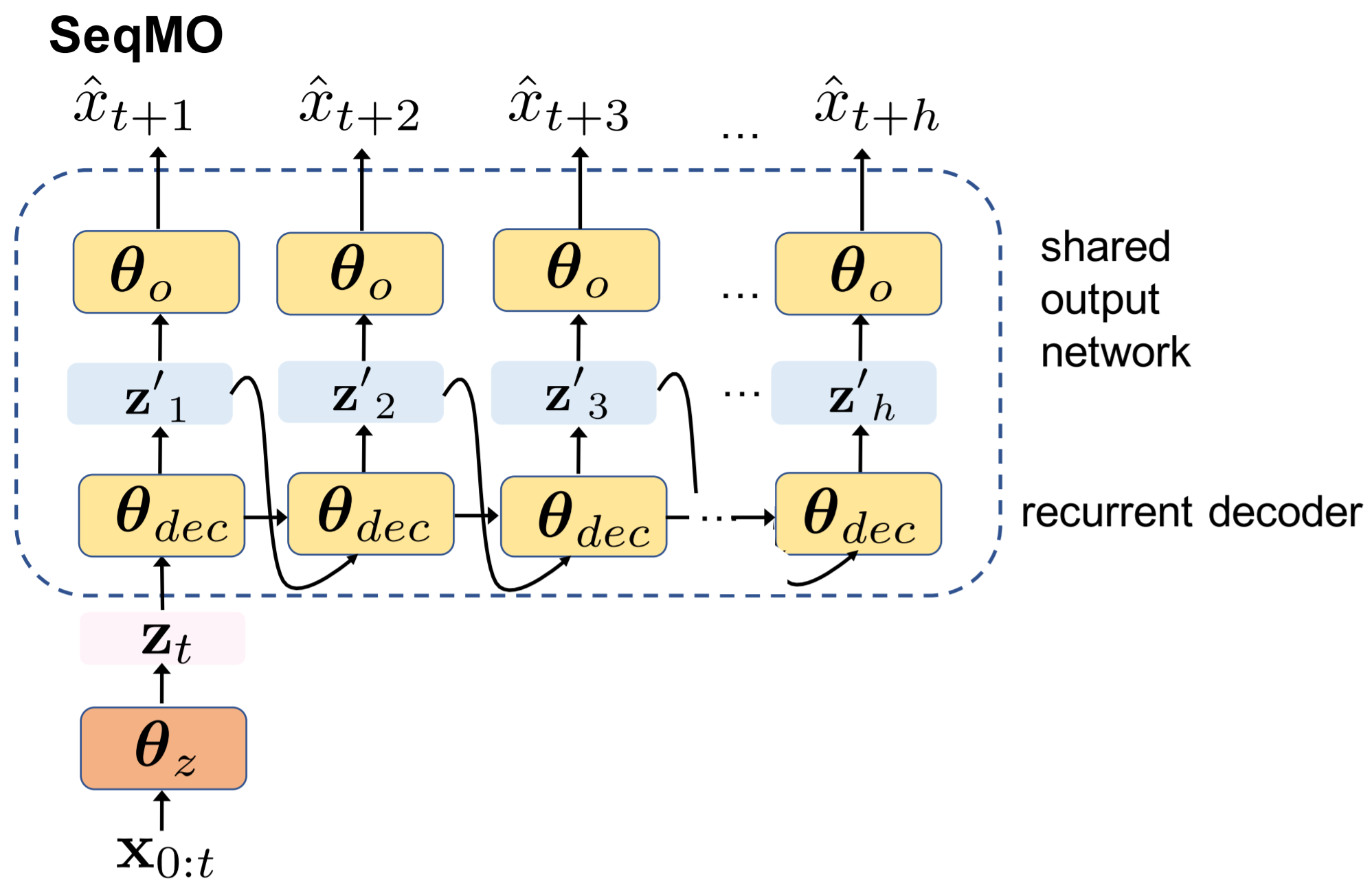}
  \caption{} \label{fig:seqmo}
\end{subfigure}
\begin{subfigure}{0.49\textwidth}
  \centering
  \includegraphics[width=\linewidth]{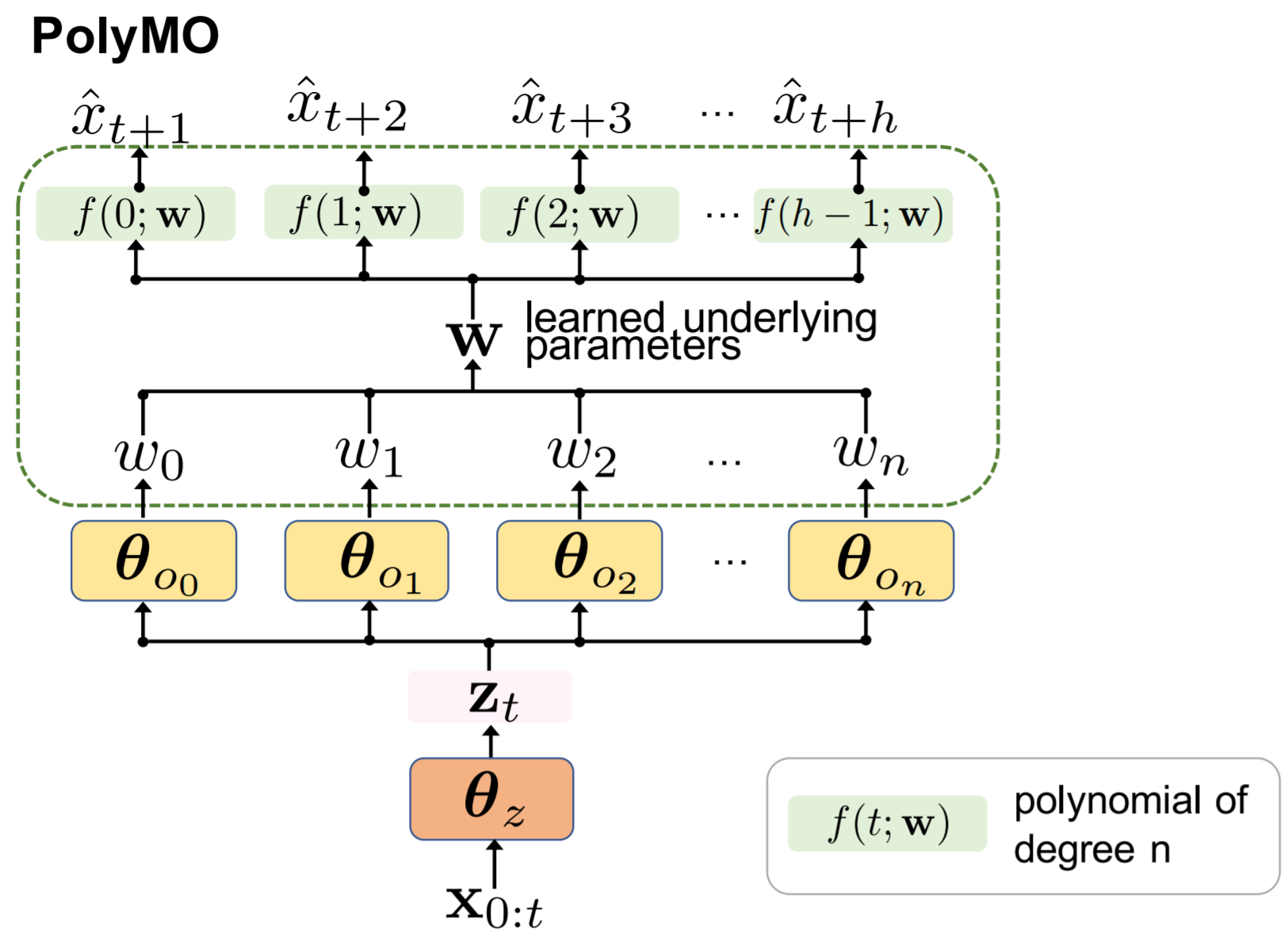}
  \caption{} \label{fig:polymo}
\end{subfigure}
\caption{Two extensions to the DeepMO forecasting framework (a) SeqMO uses a decoder network to generate a representation for each time point in the prediction horizon that feeds into a shared output network to produce $h$ predictions (b) PolyMO learns $n+1$ separate output networks based on a shared representation $z_t$, to infer the parameters of an $n^{th}$ degree polynomial which is then used to generate the predicted output.}
\label{fig:polymo_seqmo}
\end{figure*}

Our main methodological contribution is the development of a deep multi-output forecasting framework, that we extend in two directions: 1) we propose a method to propagate information across the prediction window, and 2) we propose a method to directly predict the underlying generative function of the signal. We investigate both approaches, as they represent different, complementary methods to enhance multi-output forecasting. In this section, we first describe the multi-output deep learning framework shown in \textbf{Figure \ref{fig:DeepMO}}, then explain both of our extensions, shown in \textbf{Figure \ref{fig:polymo_seqmo}}. We finish by providing additional details on how to train the models.

\subsection{Deep Multi-Output Forecasting (DeepMO)}
A neural network can function as a multi-output forecaster by using multiple output channels to infer multiple time points into the future from a shared hidden state. At time $t$, a standard multi-output neural network derives a hidden state vector $\mathbf{z}_t$ from input $\mathbf{x}_{0:t}$ through a series of hidden layers composed of linear combinations and nonlinear activations, all parameterized by $\boldsymbol{\theta}_z$. This hidden state is then translated to predictions via the network's output channels $o_{1:h}$ as follows:

\begin{gather}
\hat{x}_{t+i} = o_i(\mathbf{z}_t; \boldsymbol{\theta}_i) \text{ for } i \in [1:h] \label{eqn:output}
\end{gather}

where $o_i$ is defined in terms of linear combinations and nonlinear activations, parameterized by $\boldsymbol{\theta}_i$. This approach is illustrated in \textbf{Figure \ref{fig:DeepMO}}. Note that the value of the predicted output at time step $t+1$ is not explicitly propagated through the remainder of the prediction window (as it would be in a recursive setting). The mapping defined by $o_i$ has no direct impact on $o_j$ at inference time for $j \ne i$. However, $\hat{x}_i$ is not independent of $\hat{x}_j$, since they both depend on the shared representation $\mathbf{z}_t$. Additionally, temporal dependencies among the output are implicitly propagated by the joint optimization over $\boldsymbol{\theta} = [\boldsymbol{\theta}_z, \boldsymbol{\theta}_{o_1}, \dots\boldsymbol{\theta}_{o_h}]$ during training. However, since this approach does not explicitly encode dependencies among the outputs, learning such relationships may be more difficult. 

Standard neural networks require a fixed sized input. To eliminate this limitation, we use recurrent neural networks (RNNs), which allow for variable-sized input. This allows the network to learn the amount of history that is useful for prediction and make predictions at any point in the signal. As such, this and all subsequent architectures use recurrent cells to generate $z_t$. We use GRU cells \cite{cho_properties_2014}, however, other recurrent cells could be used as well. The recurrent cells are depicted by the orange cell in \textbf{Figure \ref{fig:DeepMO}}. We refer to the architecture described above as ``DeepMO'' (Deep Multi-Output Forecaster).

\subsection{Sequential Multi-Output Forecasting (SeqMO)}
\label{sec:seqmo} 

We extend the approach described above by combining 1) the ability of a recursive forecaster to explicitly model temporal dependencies within a sequence with 2) the ability of a multi-output system to model multiple time steps at once. To combine the advantages of these two forecasting systems, we use the DeepMO architecture, described above, but introduce temporal dependencies between sequential outputs.  

Our sequential multi-output approach modifies DeepMO by replacing the multiple output channels $o_i$ for $i\in[1:h]$ with a recurrent decoder network, parameterized by $\boldsymbol{\theta}_{dec}$. The decoder network unrolls the hidden state $\bz_t$ into $[\bz'_{1}, \bz'_{2}, \dots ,\bz'_{h}]$. Each $\bz'_{i}$ is then independently passed through the same shared output channel. Specifically, we replace (\ref{eqn:output}) with:

\begin{gather}
\bz'_{1}, \bz'_{2}, \dots,\bz'_h = Dec(\bz_t; \boldsymbol{\theta}_{dec}) \\
\hat{x}_{t+i} = o(\bz'_i; \boldsymbol{\theta}_o)
\end{gather}

With this setup, the model can learn to trade off between recursively propagating error and capturing temporal dependencies. We refer to this approach as a sequential multi-output forecaster (SeqMO) (\textbf{Figure \ref{fig:seqmo}}). We hypothesize that by explicitly encoding a temporal relationship among predictions, we will learn a more accurate forecasting strategy. This approach uses a recurrent encoding-decoding framework \cite{cho_learning_2014} for time-series forecasting. Note that this involves a many-to-many mapping, since we make multiple sequential predictions at each time step.


\subsection{Polynomial Function Forecasting (PolyMO)}
\label{sec:polymo} 
Our second proposed extension reframes the forecasting task. Instead of learning the distribution of future signal values conditioned on the past, we learn to predict an underlying representation of the data. We call this function forecasting. In particular, we assume the prediction window $\bx_{t+1:t+h} \sim f(0:h-1; \bw)$, where $\bw$ parameterizes the function class $f$. Instead of directly modeling $p(\mathbf{x}_{t+1:t+h}|\mathbf{x}_{0:t})$, we estimate the parameters to the underlying generative function $p(\bw|\mathbf{x}_{0:t})$ (see \textbf{Figure \ref{fig:polymo}}). Function forecasting is analogous to SeqMO, where input data are encoded into a hidden state best parameterizing a decoder network. The key difference is that here the decoding step is restricted to the function $f$.

We restrict our generating function class $f$ to polynomials of degree $n$. Such functions are parameterized by $n+1$ real numbers $f(t; \bw) = w_0+w_1 t + \dots + w_{n} t^n$. We modify equation (\ref{eqn:output}) so

\begin{gather}
\bw_{j} = o_j(\mathbf{z}_t; \boldsymbol{\theta}_{o_j}) \text{ for } j \in [0:n]
\end{gather}

At each time step, $t$ we predict the set of coefficients $\bw$ parameterizing the best approximation of future values $\mathbf{x}_{t+1:t+h}$. For training, we determine the actual value of the parameter by taking the best-fit polynomial of degree $n$ over $\mathbf{x}_{t+1:t+h}$. Since we want the generating function $f$ to actually model the underlying signal, and not just the observations, we limit the polynomial's capacity by setting $n << h$.

We refer to this approach of polynomial function forecasting as ``PolyMO'' (\textbf{Figure \ref{fig:polymo}}). We hypothesize that focusing on estimating the underlying generative function versus the values themselves will result in improved forecasting performance. By compactly representing future data, the output complexity of the network can be lowered, reducing noise. In addition, by predicting a generating function, the network must reason about the joint distribution of the values (since each parameter $w_i$ affects the entire output window). This helps address the error accumulation inherent to recursive forecasting. 

\subsection{Sequential Polynomial Function Forecasting (PolySeqMO)}
\label{sec:polyseqmo}
\begin{figure}[h!]
	\centering
	\includegraphics[width=0.6\linewidth]{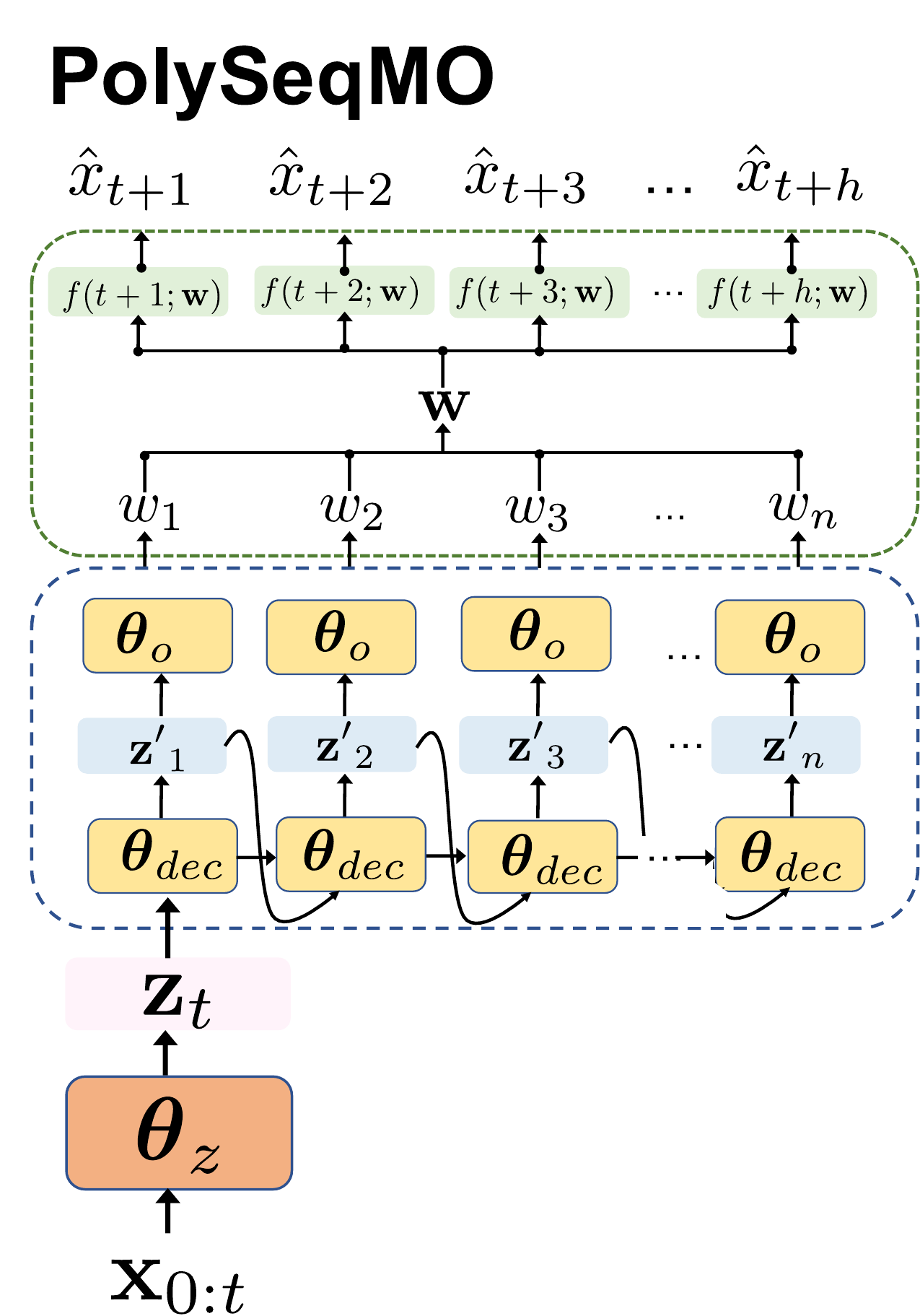}
	\caption{A combination of SeqMO and PolyMO. The function forecasting framework from PolyMO is used to predict parameters $\bw$ for a function approximating output values. These parameters are predicted using the recurrent decoding network from SeqMO.}
	\label{fig:polyseqmo}
\end{figure}
The two extensions to DeepMO shown in \textbf{Figure \ref{fig:polymo_seqmo}} both seek to improve our estimation of $p(\bx_{t+1:t+h} | \bx_{0:t})$. However, these proposed techniques represent somewhat orthogonal improvements. SeqMO provides a way to learn the trade-off between relying on intermediate value estimates and avoiding recursive error accumulation. PolyMO, meanwhile, facilitates prediction by constraining the intermediate representation, predicting values parameterizing the function approximation. The prediction of these parameters is itself a multi-output prediction. While the standard PolyMO forecaster uses the DeepMO framework to generate $w_0, \dots, w_n$, there's no reason that it couldn't use or wouldn't benefit from the SeqMO framework instead. Thus, we also examine a PolyMO forecaster with recurrent parameter decoding, denoted ``PolySeqMO'' (\textbf{Figure \ref{fig:polyseqmo}}). 

\subsection{Training and Inference Details}\label{sec:training}

In the above methods, the parameters $\boldsymbol{\theta}$ can be learned using stochastic gradient descent. The standard deep forecasting formulation defines training loss based on actual values in the signal. However, previous work has found that it can be beneficial to transform the problem into a multi-class classification task  \cite{oord_wavenet:_2016}. Thus, we replace the task of directly predicting signal values $\hat{x}_{t+i}$ with the task of predicting the probability mass function over possible discretized values of the signal: $\hat{p}(x_{t+i})$, using a cross-entropy loss against the one-hot distribution for the actual value. Each output channel $o_i$ encodes not a single number, but distribution over possible values. Similarly, we predict distributions over parameter values $\bw$ in PolyMO forecasting. While the multi-class formulation allows us to use the cross-entropy loss during training time, ultimately, we are interested in evaluating the quality of real valued forecasts. Thus, we translate these distributions to predictions by taking the value represented by the class with maximum probability in $\hat{p}$. This approach has been found to work well in the field of speech generation \cite{oord_wavenet:_2016}, but has not, to our knowledge, been investigated in the context of physiological signal forecasting.

Finally, at inference time, we smooth predictions by replacing the predicted values $\hat{\mathbf{x}}_{t+1:t+h}$ with the values occurring at that time in a best-fit polynomial, with the polynomial degree set using validation data. That is, we find the polynomial $f( \cdot; \bw)$ that best approximates $\hat{\mathbf{x}}_{t+1:t+h}$, and return as output the vector $[f(0), \dots f(h-1)]$. This polynomial smoothing allows for a more direct comparison between models that predict glucose values and the PolyMO approach.

In the sections that follow, we test our hypotheses and evaluate our proposed forecasting systems on a real dataset. We begin by describing the forecasting task, and then explain the experimental setup and provide the detailed implementation of the methods we evaluate. 

\section{Dataset \& Forecasting Task}
We consider the task for predicting future blood glucose values in patients with type 1 diabetes. These data present a challenging and clinically meaningful forecasting task. 

\subsection{The Data}
\begin{figure}[h!]
	\centering
	\includegraphics[width=1\linewidth]{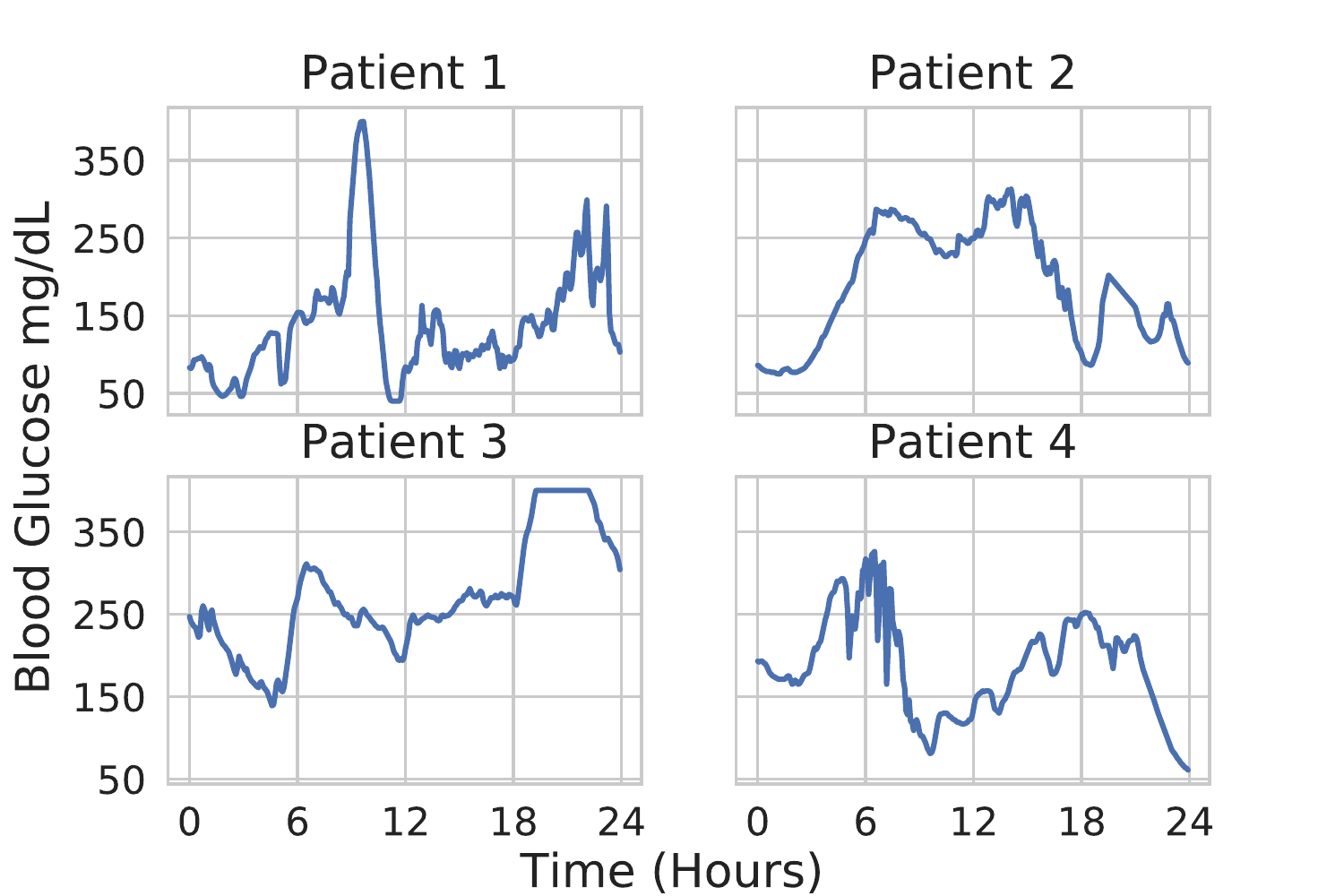}
	\caption{Four examples of continuous blood glucose values collected over the course of 24 hours from four different patients.}
	\label{fig:qual}
\end{figure}

The data consist of a large number of continuous glucose readings from 40 patients with type 1 diabetes, collected over the course of three years. At three-month intervals, individuals included in the study were given a continuous glucose monitor (CGM) that recorded their blood glucose at regular five-minute intervals over the course of several consecutive days. All subjects were blinded to the output of the CGM, so as not to affect the management of their disease. Subjects continued regular activities and insulin administration, either with injections or an insulin pump. In total, the dataset consists of 1.9k days of blood glucose measurements, totaling nearly 550k distinct glucose measurements. Blood glucose measurements were of integer resolution in the range of 40-400 mg/dL. An example of a few days worth of measurements from four different patients is found in \textbf{Figure \ref{fig:qual}}.

\subsection{The Task}
There has been extensive work on using CGM data to predict short-term outcomes (\textit{e.g.}, predicting hypo and hyperglycemic events, \cite{sudharsan_hypoglycemia_2014, oviedo_review_2017}). In contrast, here, we focus on the more general task of glucose forecasting. More specifically, we consider the challenging task of predicting future blood glucose levels using \textit{only} data about past blood glucose measurements. In this context, previous work has focused on using ARIMA \cite{eren-oruklu_hypoglycemia_2010} and machine learning algorithms such as Random Forests \cite{sudharsan_hypoglycemia_2014}. Relatedly, others have proposed machine learning techniques for leveraging data pertaining to external factors \cite{cartwright_jump_2015,plis_machine_2014,turksoy_hypoglycemia_2013}. While blood glucose is affected by external factors, \textit{e.g.,} carbohydrate intake and insulin, such data are not always readily available \cite{zecchin_how_2016} and come at the cost of increasing patient burden in the form of data collection. Through the dataset we consider does not contain these additional data, it is important to note that the proposed methods generalize to a multivariate setting. 

With enough advanced warning via a forecast, one can correct blood glucose levels through the administration of either insulin or glucose. How far in advance is far enough? It is important to note that there is 1) often a delay before insulin or glucose begins to act on the glucoregulatory system and 2) a lag between changes in blood glucose levels and CGM measurements. Thus, to have the greatest impact (\textit{e.g.}, help patients avoid hypo- and hyperglycemic events) we must be able to predict several measurements in the future. Previous work in blood glucose forecasting has settled on a 30-minute prediction window as adequate for this task \cite{cartwright_jump_2015,plis_machine_2014,turksoy_hypoglycemia_2013}. Thus, to test the efficacy of our forecasting systems, we evaluate a multi-step forecast with a 30-minute ($h=6$) prediction window.

We evaluate performance for any given prediction by calculating the mean absolute percentage error (APE) over the prediction window. This evaluation metric varies from previous work, which reports performance on only the last sample in the prediction window. We report over the entire prediction window for two reasons. First, from a clinical perspective, we are interested in the trend the values suggest. An ultimate decrease in blood glucose can have different interpretations if the rate of decline is accelerating \textit{vs.} decelerating. Second, from a technical perspective, we are interested in evaluating our systems as multi-step forecasters, naturally suggesting evaluation over multiple steps.

\section{Experimental Setup \& Baselines}
Through a series of experiments on the data and task described above, we measure the ability of our proposed methods to forecast blood glucose values. We compare against several baselines that have been used for forecasting in previous work \cite{sudharsan_hypoglycemia_2014}. We also investigate the advantage the extra supervision inherent in multi-output forecasting offers over single-output forecasting on a single-output task. 

\subsection{Train, Test, and Validation}
\begin{figure}
	\centering
	\includegraphics[width=\linewidth]{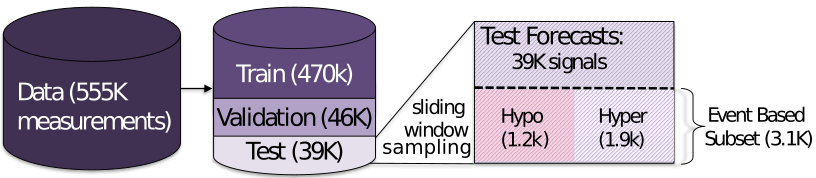}
	\caption{The splitting procedure used to train and test our models. The complete dataset is separated into three disjoint subsets: train, validation, and test. The test set is then further split into 4 non-disjoint sets: the full test set, test set examples that denote new hypo/hyperglycemic events, and subsets for each event type separately. This results in widely varying subset sizes given in the image.}
	\label{fig:train_test_split}
\end{figure}

In all of our experiments, we split the data into training, validation, and a held-out test dataset. This procedure is shown in \textbf{Figure \ref{fig:train_test_split}}. These splits were determined using the CGM recording sessions across patients. For each subject, the entirety of the final recording session is added to the test set, the second to last session is added to the validation set, and the remaining data are added to the training set. Recording sessions vary in length, but this split results in approximately 85\% of the data being used for training, 7.5\% for validation, and 7.5\% for testing. Compared to a random split, a temporal split more closely mimics how we expect the model to perform in practice. Note that several months elapse between recording sessions, so data in the training set have no immediate connection to the testing data. 

We evaluate the models at any point in time in which we have at least ten samples (\textit{i.e.}, 50 minutes) of prior data. We select this minimum to ensure that there is sufficient information to make a reasonable prediction. As CGMs are used continuously in the real world, this does not restrict applicability. We remove measurements that represent physiologically unrealistic glucose fluctuations (over 40 mg/dL in under 5 minutes) to remove noisy CGM measurements. Using a sliding window sampling method with a stride of 1 results in over 39k distinct test samples. 

For evaluation purposes, we divide our test set into four overlapping groups: 1) the full test set, 2) windows in the test set that contain either hypoglycemic onsets or hyperglycemic onsets, and two sets containing 3) only hypo and 4) only hyperglycemic onsets. Specifically, we examine performance on a second test set of samples filtered such that a hypo or hyperglycemic event begins in the 30-minute prediction window, and the last training sample was at a normal blood glucose level (between 70-180 mg/dl). Focusing on only hypo and hyperglycemic events reduces our test set size to 3,068 samples: 1,156 hypoglycemic events and 1,912 hyperglycemic events. We look at each of these subgroups to better understand forecaster performance across a range of relevant situations. The dynamics of the glucoregulatory system are highly nonlinear, and the dynamics can vary dramatically depending on the state of the glucoregulatory system and environmental contexts \cite{man_uva/padova_2014}. 

The complete test set is most representative of general model performance. However, the event test set is indicative of performance at points critical for maintaining healthy glucose levels. This event test set is further broken into a hypoglycemic event set and hyperglycemic event set. The prevention of hypo and hyperglycemic events are important for different reasons, and depending on the outcome of interest and the patient's personal history, it may be more important to effectively predict one versus the other. Thus, performance across all the test sets can be relevant. 

\subsection{Baseline Forecasting Methods}
To compare the performance of our proposed approaches to existing methods, we consider the following shallow and deep architectures.

\begin{itemize}
	\item \textbf{Baseline: Linear Extrapolation.} This baseline simply uses the most recent 30 minutes of data to extrapolate 30 minutes into the future. We chose to use the most recent 30 minutes (as opposed to a longer history) based on performance on the validation set. We include this naive baseline to give the reader a sense for how challenging the task is. It also provides an interesting comparison to the performance of the $1^{st}$ degree PolyMO and PolySeqMO models, as they have identical output capacity. 
	\item \textbf{Baseline: Random Forest (RF).} Simple but effective, the RF algorithm has been successfully used in glucose forecasting \cite{sudharsan_hypoglycemia_2014}. A robust ensemble method, it can be parallelized for rapid training and prediction. Applied to the task of predicting hypoglycemic events, Sudharsan \textit{et al.} achieved results competitive with state-of-the-art. We experimented with two random forest baselines: i) a random forest trained to predict one time step into the future, used to recursively generated the multi-output prediction, and ii) a true multi-output random forest.
	\item \textbf{Recursive RNN.} As our next baseline, we consider a recurrent neural network (RNN) which makes multi-step predictions using the recursive approach outlined in \textbf{Figure \ref{fig:new_rec}}. RNNs have recently been shown to achieve state-of-the-art results in glucose forecasting \cite{mirshekarian_using_nodate}. Our RNN uses two layers of GRU cells regularized via early stopping on a validation set and weight decay. While we are interested in minimizing the APE of our forecasts, we do not use APE as our loss function. Instead, as discussed in \textbf{Section \ref{sec:training}}, our network outputs a probability mass function over a discretized set of glucose values, thus we use a cross-entropy loss.
\end{itemize}

\subsection{Implementation Details}\label{sec:implement}
We implemented all deep learning models using PyTorch. We learned the model parameters using stochastic gradient descent with an ADAM optimizer \cite{kingma_adam:_2014}. We implemented the RF baseline using Scikit Learn \cite{pedregosa_scikit-learn:_2011}. 

All of our models have a number of hyperparameters. To ensure fair comparison between methods, we set hyperparameters by optimizing performance on the training and validation data. Our hyperparameter search space for the deep architectures included model depth, recurrent layer size, initial learning rate, and input normalization. For the RF, we tuned the number of trees and size of input.

Values reported for the RF forecaster were obtained using 100 estimators with a 10-sample input size. The remaining hyperparameters used the default Scikit Learn values. The deep architectures were found to have performance robust to hyperparameter selection. All results reported were obtained using two recurrent layers of 512 GRU hidden units. Output channels were implemented as fully connected layers with softmax activations. Training was run until performance on a separate validation set failed to increase for 50 epochs. A weight decay value of $10^{-5}$ was used for all models. All remaining model details, such as the initialization procedure and the initial learning rate for ADAM, used the PyTorch default values.

To train our PolyMO model, we tested a variety of different polynomial degrees. On the training data, we observed that blood glucose values over a 30-minute window can be well approximated with something as simple as a $1^{st}$ degree polynomial ($n=1$). On average, a linear approximation of the output window incurred a reasonably small reconstruction loss. The first degree polynomial struck a good balance between performance and capacity. On the validation data, we investigated the performance of the PolyMO model using different degrees, and found the $1^{st}$ degree performed best. 

We found the best fit $1^{st}$ degree polynomials over all length six prediction windows in the training set and used the maximum and minimum values for each coefficient as the range for our categorical output prediction, except for the bias term which we limited between 40-400 to mirror the glucose monitor output limitations. Outputs were quantized into 361 equal bins both when predicting glucose and for each polynomial coefficient. This number was chosen to give the real-value network the capacity to predict any recorded value of blood glucose, as most continuous glucose monitors have integer resolution. All source code for this project is available online \footnote{\texttt{https://github.com/igfox/multi-output-glucose-forecasting}}.

\section{Results \& Discussion}
\label{sec:res}
In total, we tested eight distinct forecasting systems: 1) Linear Extrapolation, 2) a recursive RF (RF: Rec), 3) a multi-output RF (RF: MO), 4) a recursive RNN (Recursive), 5) a multi-output RNN (DeepMO), 6) a sequential multi-output RNN (SeqMO), 7) a polynomial multi-output RNN (PolyMO), and 8) a sequential polynomial multi-output RNN (PolySeqMO). 

\textbf{Table 1} presents the forecasting model performance, in terms of APE over the prediction window in the held-out test data. We noted the error distribution was non-normal, so we report the median APE and the $2.5^{\text{th}}-97.5^{\text{th}}$ percentile errors. That said, all observed trends hold when instead considering the mean APE, with the exception that SeqMO outperforms PolySeqMO on the Event and Hypo subtasks (12.63 \textit{vs.} 12.79 and 16.60 \textit{vs.} 17.04 respectively). These results illustrate the strengths (and weaknesses) of the proposed forecasting systems applied to the task of predicting blood glucose. We discuss the implications of these results in the sections that follow. 

\begin{table*}[t!]
	\centering
	\caption{Results. We examine the performance of our eight forecasting approaches across different subsets of the CGM test data. Results are reported as 50th percentile APE over the prediction window, values in parentheses are $2.5^{\text{th}}-97.5^{\text{th}}$ percentiles. Underlined results indicate the best single-model performance. Bold results demonstrate the best overall (single or ensembled) performance.}
	\begin{tabular}{cccccc}
		\toprule
		& Architecture & \shortstack{Full\\ 39k} & \shortstack{Event\\ 3.1k} & \shortstack{Hypo\\ 1.2k} & \shortstack{Hyper\\ 1.9k} \\
		\midrule
		\multirow{3}{*}{\rotatebox[origin=c]{90}{\parbox[c]{1.5cm}{\centering Shallow Baseline}}} & &&&&\\
		& Extrapolation & 6.48 (0.21-42.12) & 10.76 (1.42-63.98) & 14.85 (1.89-86.81) & 8.73 (1.30-36.87)\\
		& RF: Rec  & 8.00 (0.62-40.83) & 10.45 (1.99-65.21) & 14.31 (2.73-91.12) & 8.82 (1.85-30.42)\\
		& RF: MO & 5.18 (0.71-30.16) & 10.64 (1.41-55.28) & 17.88 (2.70-75.46) & 8.20 (1.14-28.00)\\
		\midrule
		\multirow{3}{*}{\rotatebox[origin=c]{90}{\parbox[c]{1.2cm}{\centering Deep Baseline}}} & &&&&\\
		& Recursive & 5.31 (0.00-29.32) & 10.00 (1.45-46.22) & 13.34 (2.17-62.86) & 8.43 (1.24-30.49)\\
		& DeepMO & 5.01 (0.00-28.74) & 9.93 (1.62-41.67) & 12.91 (2.26-56.04) & 8.52 (1.43-30.02)\\
		\midrule
		\multirow{3}{*}{\rotatebox[origin=c]{90}{\parbox[c]{1.5cm}{\centering Proposed}}} & &&&&\\
		& SeqMO  & 4.91 (0.00-28.95) & 9.69 (1.51-41.54) & 12.48 (2.28-54.02) & 8.37 (1.29-29.46)\\
		& PolyMO  & 4.95 (0.51-28.30) & 9.79 (1.48-43.67) & 12.49 (1.93-60.78) & 8.46 (1.31-30.75)\\
		& PolySeqMO  & \underline{4.87} (0.48-27.80) & \underline{9.57} (1.43-43.59) & \underline{12.05} (2.03-60.90) & \underline{8.31} (1.24-29.76)\\
		& PolySeqMO Ensemble & \textbf{4.59} (0.41-21.12) & \textbf{9.38} (1.35-42.34) & \textbf{11.61} (1.99-59.89) & \textbf{8.13} (1.18-29.49)\\
		\bottomrule
	\end{tabular}
\end{table*} 

\subsection{Deep \textit{vs.} Shallow.}\label{sec:dvs} 
While RF: MO achieves good performance on the full test set, it does worse than our three improved deep multi-output methods. Moreover, it underperforms the deep approaches on the event subset. This is due mainly to very poor performance on hypoglycemic predictions. These results suggest that, compared to shallow models, deep models can more accurately learn the underlying dynamics of the glucoregulatory system from raw data. Still, we note that RF is a competitive forecaster, in line with previous work \cite{sudharsan_hypoglycemia_2014}. In particular, it achieves lower APE on the hyperglycemic test set than all models except the PolySeqMO Ensemble.

\subsection{Multi-Output \textit{vs.} Recursive.}
We observe that among both deep (DeepMO \textit{vs.} Recursive: 5.01 \textit{vs.} 5.31) and shallow approaches (RF: MO \textit{vs.} Rec: 5.18 \textit{vs.} 8.00), multi-output forecasting offers significant advantages over recursive forecasting. We also observe that all deep multi-output models improve on the Recursive model in the hypoglycemic task (12.05-12.91 \textit{vs.} 13.34). 

\begin{figure}
	\centering
	\includegraphics[width=\linewidth]{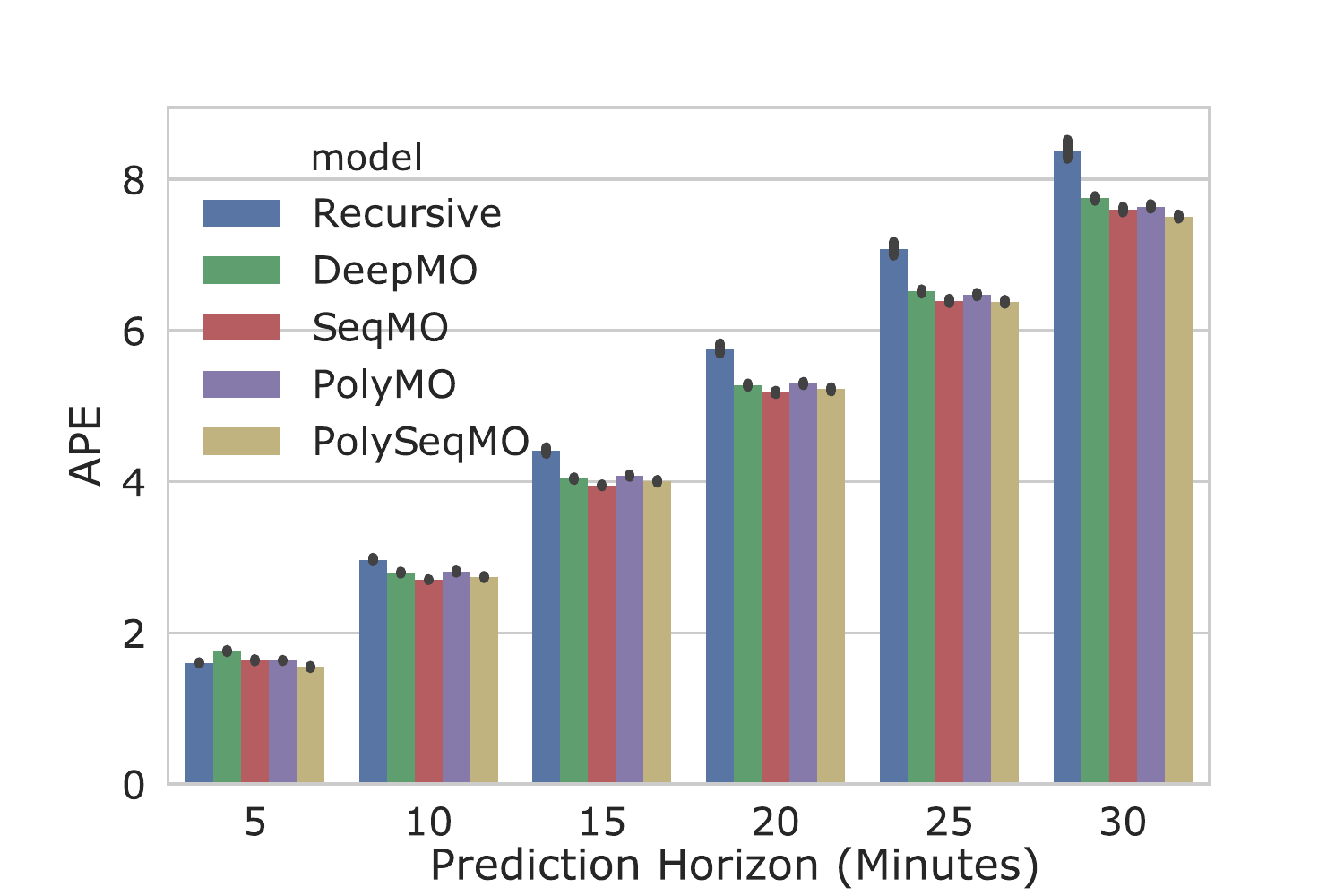}
	\caption{A comparison of per-step error between the various forecasters. While the multi-output models initially perform worse, they do not accumulate error as rapidly as the recursive approach, achieving lower error at later prediction steps.}
	\label{fig:single_vs_multi}
\end{figure}

We highlight these differences further in \textbf{Figure \ref{fig:single_vs_multi}}. We plot the average performance at each of the six time points within the 30-minute prediction window. The difference in the approaches is amplified as we predict further out. At the first place in the prediction window, corresponding to predicting $x_{t+1}$, the recursive approach outperforms most other approaches. As the target moves further in the future, we observe two trends. First, the problem becomes more difficult for all approaches (\textit{i.e.,} MO Error increases). This makes sense, as it is inherently more difficult to predict events further in the future. Second, the relative performance of the Recursive forecaster degrades with respect to the multi-output approaches. By the final step, the recursive model is far and away the worst predictor (8.38 \textit{vs.} 7.51-7.74). 

\subsection{Adding Sequential Dependencies}
Examining the difference in performance between the DeepMO and SeqMO models, we note the autoregressive connections improve performance across every subset of the data (4.91 \textit{vs.} 5.01 on the full test set). While the multi-output approach under-performs the deep recursive forecaster on the hyperglycemic task, once we add the sequential decoding, the resulting model beats the recursive forecaster on every task. This indicates that, while multi-output forecasting represents a step in the right direction, it is important to consider sequential dependencies between outputs.

\subsection{Predicting Underlying Function \textit{vs.} Values.}
\begin{figure}[t]
	\centering
	\includegraphics[width=0.7\linewidth]{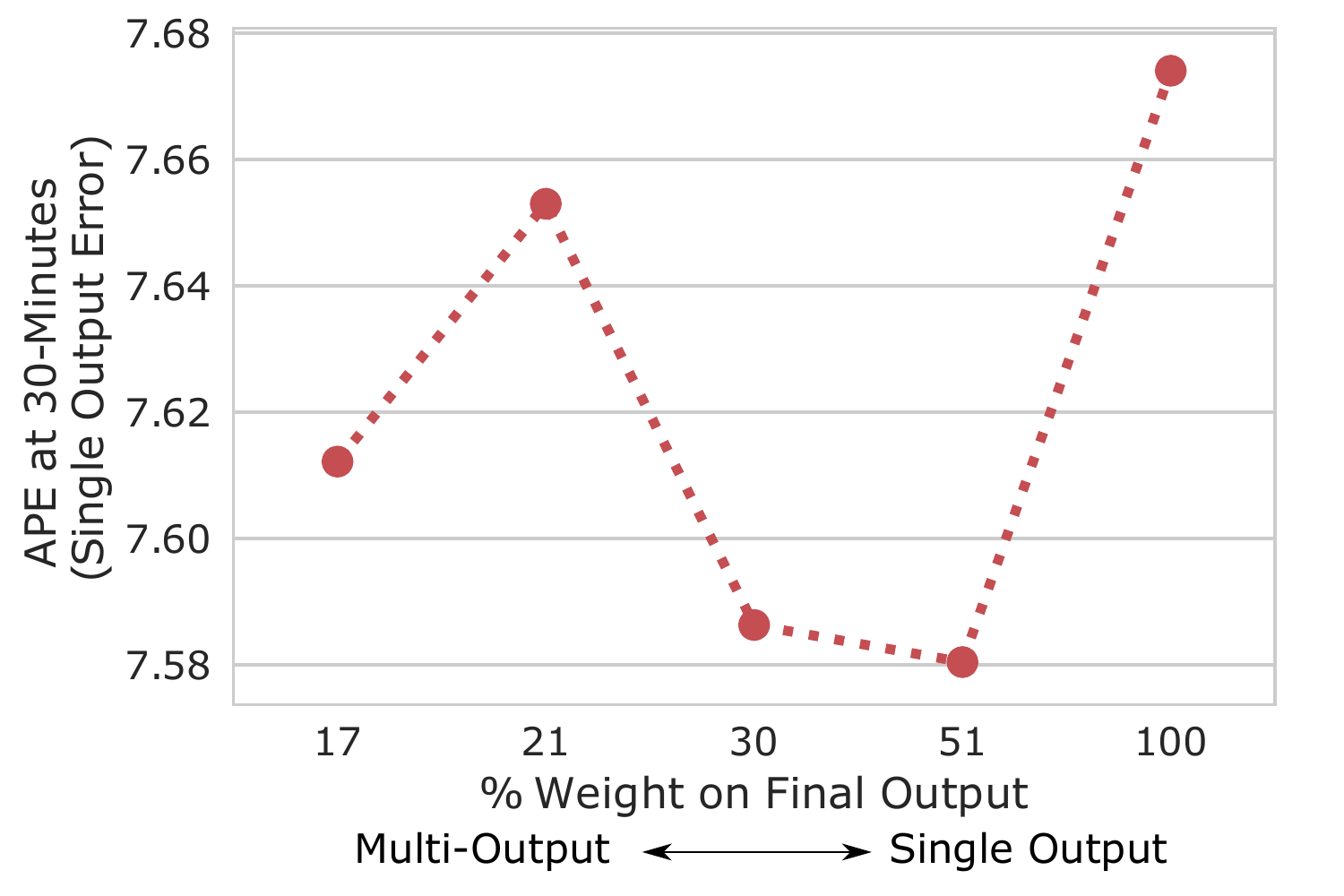}
	\caption{We examine the single output error across a range of different model types, determined using an exponential loss weighting. We derive the proportion of weight allocated to the final output (which represents the evaluation target). Surprisingly, we observe that a multi-output loss improves \textit{single}-output performance, suggesting that it is helpful to model forecast trajectories even when you only care about the final value.}
	\label{fig:mo2dir}
\end{figure}
In our investigation of PolyMO, we began by looking at the performance attained using a range of polynomial function classes. We looked at four different degree settings for the best-fit polynomial we predict (0-3 degree polynomials). We found the degree 1 model achieved the best performance on the validation set (4.87 \textit{vs.} next best 5.14). While higher degree polynomials allow for strictly better approximations of prediction windows, they also allow for more variation in output. Moreover, minor errors in high-degree coefficients rapidly compound to large errors.

Focusing on the $1^{st}$ degree PolyMO, we see it is advantageous to rephrase the value forecasting problem as a function forecasting one. In particular, we find that PolyMO beats DeepMO on every task.

Moreover, we find that the improvements from function forecasting are complimentary to those achieved by accounting for sequential output dependencies. By combining the SeqMO output decoder with the PolyMO function forecasting, resulting in PolySeqMO, we achieve better performance than all other non-ensemble models in every task under consideration. The fact that PolySeqMO does well across all subsets of the data suggests that it encodes a more accurate and complete view of the underlying dynamics of the glucoregulatory system. 

Both PolyMO and PolySeqMO focus entirely on modeling value trajectories as opposed to the values themselves. Given that we are evaluating using a multi-output metric, this built-in assumption may appear to drive the boost in performance. However, upon inspecting the performance of the Linear Extrapolation baseline, we conclude that this is not the case. Degree 1 PolyMO and PolySeqMO are equivalent in output capacity to the Linear Extrapolation baseline, and both inherently emphasize trajectories. However, the Linear Extrapolation does far worse. PolySeqMO significantly reduces the error of the Linear Extrapolation approach on the full dataset (4.87 \textit{vs.} 6.48). This demonstrates that the value of PolySeqMO is its ability to predict the future, not its assumption of linear trajectories. However, the fact that all models improve performance under polynomial smoothing suggests there is some value in the trajectory assumption.

\subsection{Ensembling}
While PolySeqMO is the best individual model in every task, it under-performs RF: MO on hyperglycemic prediction (8.31 \textit{vs.} 8.20). While this could be due to the fact that RF: MO is simply better suited to that task, it could also be a result of the general improvements observed when ensembling different model performances. To test this effect, we trained 10 PolySeqMO models on the same training set, varying only the random seed for initialization and training batch ordering. We then averaged the results of each models prediction on the test set by taking the mean.

We found that even this simple ensembling scheme with few models (relative to the 100 model ensemble used in RF: MO) leads to a sizeable increase in performance across all tasks. In particular, we find the PolySeqMO ensemble outperforms RF: MO on the hyperglycemic prediction task(8.13 \textit{vs.} 8.20).

\subsection{Multi-output \textit{vs.} Direct Forecasting}
There are many forecasting problems in which an accurate single-output forecast may suffice. In such cases, it is common to use a direct forecaster \cite{taieb_review_2012}, where one \textit{directly} estimates $p(\mathbf{x}_{t+h}|\mathbf{x}_{0:t})$. In a follow-up analysis, we demonstrate that even in cases where only a single output is desired, it can be beneficial to consider a multi-output forecasting framework.

To demonstrate this, we begin by introducing a method to transition from multi-output to direct forecasting, focusing on SeqMO. This model operates by predicting multiple values at each time step. The training loss is the average across each of the six time steps in the prediction window. A direct forecaster, aiming to make a single prediction at the final time step, can be approximated by zeroing out all losses except those incurred at the final step. This focuses the full capacity of the network on predicting the final value. We can flexibly transition between direct and multi-output forecasting by manipulating the per-step loss weighting, transitioning from a one-hot vector on the final output (direct) to a uniform allocation of weight across the window (multi-output). We encapsulate this transition in a single hyperparameter, $0 \le b_w \le 1$, or the base for the per-step loss weight. For each step $i$ in a prediction window of length $h$, we set the loss weight $w_{l,i} = \frac{b_w ^{h-i}}{\sum_{i=0}^{h}b_w ^{h-i}}$. 

In \textbf{Figure \ref{fig:mo2dir}} we show the single output performance (predicting 30 minutes into the future) of SeqMO with different settings of $b_w$. We observe that modeling the full trajectory does not worsen performance, and in fact appears to slightly improve it (MO 7.67 \textit{vs.} Direct 7.61). Interestingly, we found that best single-output performance was achieved using an intermediate value for $b_w$ (7.58 with $b_w = 0.5$). Mixing multi-output and direct forecasting strategies could be a promising direction for improving single-output forecasting performance. 

\section{Conclusions}
In this work, we investigated methods for deep multi-output blood glucose forecasting. We demonstrated the importance of balancing autoregressive behavior and sequential error accumulation, and provided a forecasting model, SeqMO, that accomplishes this. We developed the idea of function forecasting, and introduced novel forecasting methods PolyMO and PolySeqMO. We compared our proposed approaches to both shallow and deep baselines. Applied to the challenging task of predicting blood glucose, we demonstrated that: 1) multi-output methods outperform recursive alternatives, 2) modeling underlying dependencies among outputs using explicit connections and function forecasting leads to better performance, and 3) the proposed approaches are complementary, and combining them significantly improves performance. Additionally, we demonstrated that multi-output forecasting improves performance, even on a single-output task.

These experimental results suggest a multi-output approach can effectively capture the underlying dynamics of the glucoregulatory system. While we focus on blood glucose, forecasting real valued signals is a problem with applications across a number of different domains including speech processing, weather prediction, and medicine \cite{ghassemi_multivariate_2015,oord_wavenet:_2016,gensler_deep_2016}. Our proposed methods are generally applicable to any forecasting problem requiring multi-step predictions. 

\section{Acknowledgements}
This work was supported by the National Science Foundation (NSF award no. IIS-1553146), the Michigan Institute for Data Science (MIDAS), and the National Institutes of Health (NIH research grant RO1 NIH/NHLBI1R01HL102334-01). The views and conclusions in this document are those of the authors and should not be interpreted as necessarily representing the official policies, either expressed or implied, of the NSF or the NIH

\bibliographystyle{ACM-Reference-Format}
\bibliography{references}


\begin{thebibliography}{27}


\ifx \showCODEN    \undefined \def \showCODEN     #1{\unskip}     \fi
\ifx \showDOI      \undefined \def \showDOI       #1{#1}\fi
\ifx \showISBNx    \undefined \def \showISBNx     #1{\unskip}     \fi
\ifx \showISBNxiii \undefined \def \showISBNxiii  #1{\unskip}     \fi
\ifx \showISSN     \undefined \def \showISSN      #1{\unskip}     \fi
\ifx \showLCCN     \undefined \def \showLCCN      #1{\unskip}     \fi
\ifx \shownote     \undefined \def \shownote      #1{#1}          \fi
\ifx \showarticletitle \undefined \def \showarticletitle #1{#1}   \fi
\ifx \showURL      \undefined \def \showURL       {\relax}        \fi
\providecommand\bibfield[2]{#2}
\providecommand\bibinfo[2]{#2}
\providecommand\natexlab[1]{#1}
\providecommand\showeprint[2][]{arXiv:#2}

\bibitem[\protect\citeauthoryear{Bengio, Vinyals, Jaitly, and Shazeer}{Bengio
  et~al\mbox{.}}{2015}]%
        {bengio_scheduled_2015}
\bibfield{author}{\bibinfo{person}{Samy Bengio}, \bibinfo{person}{Oriol
  Vinyals}, \bibinfo{person}{Navdeep Jaitly}, {and} \bibinfo{person}{Noam
  Shazeer}.} \bibinfo{year}{2015}\natexlab{}.
\newblock \showarticletitle{Scheduled {Sampling} for {Sequence} {Prediction}
  with {Recurrent} {Neural} {Networks}}.
\newblock \bibinfo{journal}{\emph{arXiv:1506.03099 [cs]}} (\bibinfo{date}{June}
  \bibinfo{year}{2015}).
\newblock
\urldef\tempurl%
\url{http://arxiv.org/abs/1506.03099}
\showURL{%
\tempurl}
\newblock
\shownote{arXiv: 1506.03099.}


\bibitem[\protect\citeauthoryear{Cho, van Merrienboer, Bahdanau, and
  Bengio}{Cho et~al\mbox{.}}{2014a}]%
        {cho_properties_2014}
\bibfield{author}{\bibinfo{person}{Kyunghyun Cho}, \bibinfo{person}{Bart van
  Merrienboer}, \bibinfo{person}{Dzmitry Bahdanau}, {and}
  \bibinfo{person}{Yoshua Bengio}.} \bibinfo{year}{2014}\natexlab{a}.
\newblock \showarticletitle{On the {Properties} of {Neural} {Machine}
  {Translation}: {Encoder}-{Decoder} {Approaches}}.
\newblock \bibinfo{journal}{\emph{arXiv:1409.1259 [cs, stat]}}
  (\bibinfo{date}{Sept.} \bibinfo{year}{2014}).
\newblock
\newblock
\shownote{arXiv: 1409.1259.}


\bibitem[\protect\citeauthoryear{Cho, van Merrienboer, Gulcehre, Bahdanau,
  Bougares, Schwenk, and Bengio}{Cho et~al\mbox{.}}{2014b}]%
        {cho_learning_2014}
\bibfield{author}{\bibinfo{person}{Kyunghyun Cho}, \bibinfo{person}{Bart van
  Merrienboer}, \bibinfo{person}{Caglar Gulcehre}, \bibinfo{person}{Dzmitry
  Bahdanau}, \bibinfo{person}{Fethi Bougares}, \bibinfo{person}{Holger
  Schwenk}, {and} \bibinfo{person}{Yoshua Bengio}.}
  \bibinfo{year}{2014}\natexlab{b}.
\newblock \showarticletitle{Learning {Phrase} {Representations} using {RNN}
  {Encoder}-{Decoder} for {Statistical} {Machine} {Translation}}.
\newblock  (\bibinfo{date}{June} \bibinfo{year}{2014}).
\newblock
\newblock
\shownote{arXiv: 1406.1078.}


\bibitem[\protect\citeauthoryear{Cobelli, Renard, and Kovatchev}{Cobelli
  et~al\mbox{.}}{2011}]%
        {cobelli_artificial_2011}
\bibfield{author}{\bibinfo{person}{Claudio Cobelli}, \bibinfo{person}{Eric
  Renard}, {and} \bibinfo{person}{Boris Kovatchev}.}
  \bibinfo{year}{2011}\natexlab{}.
\newblock \showarticletitle{Artificial pancreas: past, present, future}.
\newblock \bibinfo{journal}{\emph{Diabetes}} \bibinfo{volume}{60},
  \bibinfo{number}{11} (\bibinfo{year}{2011}), \bibinfo{pages}{2672--2682}.
\newblock


\bibitem[\protect\citeauthoryear{Eren-Oruklu, Cinar, and Quinn}{Eren-Oruklu
  et~al\mbox{.}}{2010}]%
        {eren-oruklu_hypoglycemia_2010}
\bibfield{author}{\bibinfo{person}{Meriyan Eren-Oruklu}, \bibinfo{person}{Ali
  Cinar}, {and} \bibinfo{person}{Lauretta Quinn}.}
  \bibinfo{year}{2010}\natexlab{}.
\newblock \bibinfo{booktitle}{\emph{Hypoglycemia prediction with
  subject-specific recursive time-series models}}.
\newblock \bibinfo{publisher}{SAGE Publications}.
\newblock


\bibitem[\protect\citeauthoryear{Gensler, Henze, Sick, and Raabe}{Gensler
  et~al\mbox{.}}{2016}]%
        {gensler_deep_2016}
\bibfield{author}{\bibinfo{person}{Andre Gensler}, \bibinfo{person}{Janosch
  Henze}, \bibinfo{person}{Bernhard Sick}, {and} \bibinfo{person}{Nils Raabe}.}
  \bibinfo{year}{2016}\natexlab{}.
\newblock \showarticletitle{Deep {Learning} for solar power forecasting
  \#x2014; {An} approach using {AutoEncoder} and {LSTM} {Neural} {Networks}}.
  In \bibinfo{booktitle}{\emph{2016 {IEEE} {International} {Conference} on
  {Systems}, {Man}, and {Cybernetics} ({SMC})}}.
  \bibinfo{pages}{002858--002865}.
\newblock


\bibitem[\protect\citeauthoryear{Ghassemi, Pimentel, Naumann, Brennan, Clifton,
  Szolovits, and Feng}{Ghassemi et~al\mbox{.}}{2015}]%
        {ghassemi_multivariate_2015}
\bibfield{author}{\bibinfo{person}{Marzyeh Ghassemi}, \bibinfo{person}{Marco~AF
  Pimentel}, \bibinfo{person}{Tristan Naumann}, \bibinfo{person}{Thomas
  Brennan}, \bibinfo{person}{David~A. Clifton}, \bibinfo{person}{Peter
  Szolovits}, {and} \bibinfo{person}{Mengling Feng}.}
  \bibinfo{year}{2015}\natexlab{}.
\newblock \showarticletitle{A multivariate timeseries modeling approach to
  severity of illness assessment and forecasting in icu with sparse,
  heterogeneous clinical data}. In \bibinfo{booktitle}{\emph{Proceedings of
  the... {AAAI} {Conference} on {Artificial} {Intelligence}. {AAAI}
  {Conference} on {Artificial} {Intelligence}}}, Vol.~\bibinfo{volume}{2015}.
  \bibinfo{publisher}{NIH Public Access}, \bibinfo{pages}{446}.
\newblock


\bibitem[\protect\citeauthoryear{Group}{Group}{[n. d.]a}]%
        {noauthor_effect_1993}
\bibfield{author}{\bibinfo{person}{DCCT/EDIC~Research Group}.}
  \bibinfo{year}{[n. d.]}\natexlab{a}.
\newblock \showarticletitle{The {Effect} of {Intensive} {Treatment} of
  {Diabetes} on the {Development} and {Progression} of {Long}-{Term}
  {Complications} in {Insulin}-{Dependent} {Diabetes} {Mellitus}}.
\newblock  (\bibinfo{year}{[n. d.]}).
\newblock


\bibitem[\protect\citeauthoryear{Group}{Group}{[n. d.]b}]%
        {noauthor_intensive_2011}
\bibfield{author}{\bibinfo{person}{DCCT/EDIC~Research Group}.}
  \bibinfo{year}{[n. d.]}\natexlab{b}.
\newblock \showarticletitle{Intensive {Diabetes} {Therapy} and {Glomerular}
  {Filtration} {Rate} in {Type} 1 {Diabetes}}.
\newblock  (\bibinfo{year}{[n. d.]}).
\newblock


\bibitem[\protect\citeauthoryear{Group}{Group}{[n. d.]c}]%
  {writing_team_for_the_diabetes_control_and_complications_trial/epidemiology_of_diabetes_interventions_and_complications_research_group_sustained_2003}
\bibfield{author}{\bibinfo{person}{DCCT/EDIC~Research Group}.}
  \bibinfo{year}{[n. d.]}\natexlab{c}.
\newblock \showarticletitle{Sustained effect of intensive treatment of type 1
  diabetes mellitus on development and progression of diabetic nephropathy: the
  {Epidemiology} of {Diabetes} {Interventions} and {Complications} ({EDIC})
  study}.
\newblock  (\bibinfo{year}{[n. d.]}).
\newblock


\bibitem[\protect\citeauthoryear{Hovorka, Canonico, Chassin, Haueter,
  Massi-Benedetti, Federici, Pieber, Schaller, Schaupp, Vering, and
  {others}}{Hovorka et~al\mbox{.}}{2004}]%
        {hovorka_nonlinear_2004}
\bibfield{author}{\bibinfo{person}{Roman Hovorka}, \bibinfo{person}{Valentina
  Canonico}, \bibinfo{person}{Ludovic~J. Chassin}, \bibinfo{person}{Ulrich
  Haueter}, \bibinfo{person}{Massimo Massi-Benedetti},
  \bibinfo{person}{Marco~Orsini Federici}, \bibinfo{person}{Thomas~R. Pieber},
  \bibinfo{person}{Helga~C. Schaller}, \bibinfo{person}{Lukas Schaupp},
  \bibinfo{person}{Thomas Vering}, {and} \bibinfo{person}{{others}}.}
  \bibinfo{year}{2004}\natexlab{}.
\newblock \showarticletitle{Nonlinear model predictive control of glucose
  concentration in subjects with type 1 diabetes}.
\newblock \bibinfo{journal}{\emph{Physiological measurement}}
  \bibinfo{volume}{25}, \bibinfo{number}{4} (\bibinfo{year}{2004}),
  \bibinfo{pages}{905}.
\newblock


\bibitem[\protect\citeauthoryear{Kingma and Ba}{Kingma and Ba}{2014}]%
        {kingma_adam:_2014}
\bibfield{author}{\bibinfo{person}{Diederik Kingma} {and}
  \bibinfo{person}{Jimmy Ba}.} \bibinfo{year}{2014}\natexlab{}.
\newblock \showarticletitle{Adam: {A} method for stochastic optimization}.
\newblock \bibinfo{journal}{\emph{arXiv preprint arXiv:1412.6980}}
  (\bibinfo{year}{2014}).
\newblock


\bibitem[\protect\citeauthoryear{Lamb, Goyal, Zhang, Zhang, Courville, and
  Bengio}{Lamb et~al\mbox{.}}{2016}]%
        {lamb_professor_2016}
\bibfield{author}{\bibinfo{person}{Alex Lamb}, \bibinfo{person}{Anirudh Goyal},
  \bibinfo{person}{Ying Zhang}, \bibinfo{person}{Saizheng Zhang},
  \bibinfo{person}{Aaron Courville}, {and} \bibinfo{person}{Yoshua Bengio}.}
  \bibinfo{year}{2016}\natexlab{}.
\newblock \showarticletitle{Professor {Forcing}: {A} {New} {Algorithm} for
  {Training} {Recurrent} {Networks}}.
\newblock \bibinfo{journal}{\emph{arXiv:1610.09038 [cs, stat]}}
  (\bibinfo{date}{Oct.} \bibinfo{year}{2016}).
\newblock
\urldef\tempurl%
\url{http://arxiv.org/abs/1610.09038}
\showURL{%
\tempurl}
\newblock
\shownote{arXiv: 1610.09038.}


\bibitem[\protect\citeauthoryear{Man, Micheletto, Lv, Breton, Kovatchev, and
  Cobelli}{Man et~al\mbox{.}}{2014}]%
        {man_uva/padova_2014}
\bibfield{author}{\bibinfo{person}{Chiara~Dalla Man},
  \bibinfo{person}{Francesco Micheletto}, \bibinfo{person}{Dayu Lv},
  \bibinfo{person}{Marc Breton}, \bibinfo{person}{Boris Kovatchev}, {and}
  \bibinfo{person}{Claudio Cobelli}.} \bibinfo{year}{2014}\natexlab{}.
\newblock \showarticletitle{The {UVA}/{PADOVA} type 1 diabetes simulator: new
  features}.
\newblock \bibinfo{journal}{\emph{Journal of diabetes science and technology}}
  \bibinfo{volume}{8}, \bibinfo{number}{1} (\bibinfo{year}{2014}),
  \bibinfo{pages}{26--34}.
\newblock


\bibitem[\protect\citeauthoryear{Mirshekarian, Bunescu, Marling, and
  Schwartz}{Mirshekarian et~al\mbox{.}}{2017}]%
        {mirshekarian_using_nodate}
\bibfield{author}{\bibinfo{person}{Sadegh Mirshekarian},
  \bibinfo{person}{Razvan Bunescu}, \bibinfo{person}{Cindy Marling}, {and}
  \bibinfo{person}{Frank Schwartz}.} \bibinfo{year}{2017}\natexlab{}.
\newblock \showarticletitle{Using {LSTMs} to {Learn} {Physiological} {Models}
  of {Blood} {Glucose} {Behavior}}.
\newblock \bibinfo{journal}{\emph{EMBC}} (\bibinfo{year}{2017}).
\newblock


\bibitem[\protect\citeauthoryear{Nathan, Bayless, Cleary, Genuth,
  Gubitosi-Klug, Lachin, Lorenzi, Zinman, and Group}{Nathan
  et~al\mbox{.}}{2013}]%
        {nathan_diabetes_2013}
\bibfield{author}{\bibinfo{person}{David~M. Nathan}, \bibinfo{person}{Margaret
  Bayless}, \bibinfo{person}{Patricia Cleary}, \bibinfo{person}{Saul Genuth},
  \bibinfo{person}{Rose Gubitosi-Klug}, \bibinfo{person}{John~M. Lachin},
  \bibinfo{person}{Gayle Lorenzi}, \bibinfo{person}{Bernard Zinman}, {and}
  \bibinfo{person}{for the DCCT/EDIC~Research Group}.}
  \bibinfo{year}{2013}\natexlab{}.
\newblock \showarticletitle{Diabetes {Control} and {Complications}
  {Trial}/{Epidemiology} of {Diabetes} {Interventions} and {Complications}
  {Study} at 30 {Years}: {Advances} and {Contributions}}.
\newblock \bibinfo{journal}{\emph{Diabetes}} \bibinfo{volume}{62},
  \bibinfo{number}{12} (\bibinfo{date}{Dec.} \bibinfo{year}{2013}),
  \bibinfo{pages}{3976--3986}.
\newblock
\showISSN{0012-1797, 1939-327X}
\urldef\tempurl%
\url{https://doi.org/10.2337/db13-1093}
\showDOI{\tempurl}


\bibitem[\protect\citeauthoryear{Oord, Dieleman, Zen, Simonyan, Vinyals,
  Graves, Kalchbrenner, Senior, and Kavukcuoglu}{Oord et~al\mbox{.}}{2016}]%
        {oord_wavenet:_2016}
\bibfield{author}{\bibinfo{person}{Aaron van~den Oord}, \bibinfo{person}{Sander
  Dieleman}, \bibinfo{person}{Heiga Zen}, \bibinfo{person}{Karen Simonyan},
  \bibinfo{person}{Oriol Vinyals}, \bibinfo{person}{Alex Graves},
  \bibinfo{person}{Nal Kalchbrenner}, \bibinfo{person}{Andrew Senior}, {and}
  \bibinfo{person}{Koray Kavukcuoglu}.} \bibinfo{year}{2016}\natexlab{}.
\newblock \showarticletitle{Wavenet: {A} generative model for raw audio}.
\newblock \bibinfo{journal}{\emph{arXiv preprint arXiv:1609.03499}}
  (\bibinfo{year}{2016}).
\newblock


\bibitem[\protect\citeauthoryear{Oviedo, Vehí, Calm, and Armengol}{Oviedo
  et~al\mbox{.}}{2017}]%
        {oviedo_review_2017}
\bibfield{author}{\bibinfo{person}{Silvia Oviedo}, \bibinfo{person}{Josep
  Vehí}, \bibinfo{person}{Remei Calm}, {and} \bibinfo{person}{Joaquim
  Armengol}.} \bibinfo{year}{2017}\natexlab{}.
\newblock \showarticletitle{A review of personalized blood glucose prediction
  strategies for {T}1DM patients}.
\newblock \bibinfo{journal}{\emph{International journal for numerical methods
  in biomedical engineering}} \bibinfo{volume}{33}, \bibinfo{number}{6}
  (\bibinfo{year}{2017}).
\newblock


\bibitem[\protect\citeauthoryear{Pedregosa, Varoquaux, Gramfort, Michel,
  Thirion, Grisel, Blondel, Prettenhofer, Weiss, Dubourg, and
  {others}}{Pedregosa et~al\mbox{.}}{2011}]%
        {pedregosa_scikit-learn:_2011}
\bibfield{author}{\bibinfo{person}{Fabian Pedregosa}, \bibinfo{person}{Gaël
  Varoquaux}, \bibinfo{person}{Alexandre Gramfort}, \bibinfo{person}{Vincent
  Michel}, \bibinfo{person}{Bertrand Thirion}, \bibinfo{person}{Olivier
  Grisel}, \bibinfo{person}{Mathieu Blondel}, \bibinfo{person}{Peter
  Prettenhofer}, \bibinfo{person}{Ron Weiss}, \bibinfo{person}{Vincent
  Dubourg}, {and} \bibinfo{person}{{others}}.} \bibinfo{year}{2011}\natexlab{}.
\newblock \showarticletitle{Scikit-learn: {Machine} learning in {Python}}.
\newblock \bibinfo{journal}{\emph{Journal of Machine Learning Research}}
  \bibinfo{volume}{12}, \bibinfo{number}{Oct} (\bibinfo{year}{2011}),
  \bibinfo{pages}{2825--2830}.
\newblock


\bibitem[\protect\citeauthoryear{Plis, Bunescu, Marling, Shubrook, and
  Schwartz}{Plis et~al\mbox{.}}{2014}]%
        {plis_machine_2014}
\bibfield{author}{\bibinfo{person}{Kevin Plis}, \bibinfo{person}{Razvan
  Bunescu}, \bibinfo{person}{Cindy Marling}, \bibinfo{person}{Jay Shubrook},
  {and} \bibinfo{person}{Frank Schwartz}.} \bibinfo{year}{2014}\natexlab{}.
\newblock \showarticletitle{A machine learning approach to predicting blood
  glucose levels for diabetes management}.
\newblock \bibinfo{journal}{\emph{Modern Artificial Intelligence for Health
  Analytics. Papers from the AAAI-14}} (\bibinfo{year}{2014}).
\newblock


\bibitem[\protect\citeauthoryear{Sudharsan, Peeples, and Shomali}{Sudharsan
  et~al\mbox{.}}{2014}]%
        {sudharsan_hypoglycemia_2014}
\bibfield{author}{\bibinfo{person}{Bharath Sudharsan}, \bibinfo{person}{Malinda
  Peeples}, {and} \bibinfo{person}{Mansur Shomali}.}
  \bibinfo{year}{2014}\natexlab{}.
\newblock \showarticletitle{Hypoglycemia prediction using machine learning
  models for patients with type 2 diabetes}.
\newblock \bibinfo{journal}{\emph{Journal of diabetes science and technology}}
  \bibinfo{volume}{9}, \bibinfo{number}{1} (\bibinfo{year}{2014}),
  \bibinfo{pages}{86--90}.
\newblock


\bibitem[\protect\citeauthoryear{Taieb, Bontempi, Atiya, and Sorjamaa}{Taieb
  et~al\mbox{.}}{2012}]%
        {taieb_review_2012}
\bibfield{author}{\bibinfo{person}{Souhaib~Ben Taieb},
  \bibinfo{person}{Gianluca Bontempi}, \bibinfo{person}{Amir Atiya}, {and}
  \bibinfo{person}{Antti Sorjamaa}.} \bibinfo{year}{2012}\natexlab{}.
\newblock \showarticletitle{A review and comparison of strategies for
  multi-step ahead time series forecasting based on the {NN}5 forecasting
  competition}.
\newblock \bibinfo{journal}{\emph{Expert systems with applications}}
  \bibinfo{volume}{39}, \bibinfo{number}{8} (\bibinfo{year}{2012}),
  \bibinfo{pages}{7067--7083}.
\newblock


\bibitem[\protect\citeauthoryear{Tao, Pietropaolo, Atkinson, Schatz, and
  Taylor}{Tao et~al\mbox{.}}{2010}]%
        {tao_estimating_2010}
\bibfield{author}{\bibinfo{person}{Betty Tao}, \bibinfo{person}{Massimo
  Pietropaolo}, \bibinfo{person}{Mark Atkinson}, \bibinfo{person}{Desmond
  Schatz}, {and} \bibinfo{person}{David Taylor}.}
  \bibinfo{year}{2010}\natexlab{}.
\newblock \showarticletitle{Estimating the {Cost} of {Type} 1 {Diabetes} in the
  {U}.{S}.: {A} {Propensity} {Score} {Matching} {Method}}.
\newblock \bibinfo{journal}{\emph{PLOS ONE}} \bibinfo{volume}{5},
  \bibinfo{number}{7} (\bibinfo{date}{July} \bibinfo{year}{2010}),
  \bibinfo{pages}{e11501}.
\newblock
\showISSN{1932-6203}


\bibitem[\protect\citeauthoryear{Turksoy, Bayrak, Quinn, Littlejohn, Rollins,
  and Cinar}{Turksoy et~al\mbox{.}}{2013}]%
        {turksoy_hypoglycemia_2013}
\bibfield{author}{\bibinfo{person}{Kamuran Turksoy}, \bibinfo{person}{Elif~S.
  Bayrak}, \bibinfo{person}{Lauretta Quinn}, \bibinfo{person}{Elizabeth
  Littlejohn}, \bibinfo{person}{Derrick Rollins}, {and} \bibinfo{person}{Ali
  Cinar}.} \bibinfo{year}{2013}\natexlab{}.
\newblock \showarticletitle{Hypoglycemia early alarm systems based on
  multivariable models}.
\newblock \bibinfo{journal}{\emph{Industrial \& engineering chemistry
  research}} \bibinfo{volume}{52}, \bibinfo{number}{35} (\bibinfo{year}{2013}),
  \bibinfo{pages}{12329--12336}.
\newblock


\bibitem[\protect\citeauthoryear{Wu, Chen, Qiao, and Lu}{Wu
  et~al\mbox{.}}{2016}]%
        {wu_probabilistic_2016}
\bibfield{author}{\bibinfo{person}{Wenzu Wu}, \bibinfo{person}{Kunjin Chen},
  \bibinfo{person}{Ying Qiao}, {and} \bibinfo{person}{Zongxiang Lu}.}
  \bibinfo{year}{2016}\natexlab{}.
\newblock \showarticletitle{Probabilistic short-term wind power forecasting
  based on deep neural networks}. In \bibinfo{booktitle}{\emph{2016
  {International} {Conference} on {Probabilistic} {Methods} {Applied} to
  {Power} {Systems} ({PMAPS})}}. \bibinfo{pages}{1--8}.
\newblock


\bibitem[\protect\citeauthoryear{Zecchin, Facchinetti, Sparacino, and
  Cobelli}{Zecchin et~al\mbox{.}}{2015}]%
        {cartwright_jump_2015}
\bibfield{author}{\bibinfo{person}{Chiara Zecchin}, \bibinfo{person}{Andrea
  Facchinetti}, \bibinfo{person}{Giovanni Sparacino}, {and}
  \bibinfo{person}{Claudio Cobelli}.} \bibinfo{year}{2015}\natexlab{}.
\newblock \showarticletitle{Jump {Neural} {Network} for {Real}-{Time}
  {Prediction} of {Glucose} {Concentration}}.
\newblock In \bibinfo{booktitle}{\emph{Artificial {Neural} {Networks}}}.
  \bibinfo{publisher}{Springer New York}, \bibinfo{pages}{245--259}.
\newblock
\showISBNx{978-1-4939-2238-3}
\newblock
\shownote{DOI: 10.1007/978-1-4939-2239-0\_15.}


\bibitem[\protect\citeauthoryear{Zecchin, Facchinetti, Sparacino, and
  Cobelli}{Zecchin et~al\mbox{.}}{2016}]%
        {zecchin_how_2016}
\bibfield{author}{\bibinfo{person}{Chiara Zecchin}, \bibinfo{person}{Andrea
  Facchinetti}, \bibinfo{person}{Giovanni Sparacino}, {and}
  \bibinfo{person}{Claudio Cobelli}.} \bibinfo{year}{2016}\natexlab{}.
\newblock \showarticletitle{How {Much} {Is} {Short}-{Term} {Glucose}
  {Prediction} in {Type} 1 {Diabetes} {Improved} by {Adding} {Insulin}
  {Delivery} and {Meal} {Content} {Information} to {CGM} {Data}? {A}
  {Proof}-of-{Concept} {Study}}.
\newblock \bibinfo{journal}{\emph{Journal of Diabetes Science and Technology}}
  \bibinfo{volume}{10}, \bibinfo{number}{5} (\bibinfo{date}{Sept.}
  \bibinfo{year}{2016}), \bibinfo{pages}{1149--1160}.
\newblock
\showISSN{1932-2968}


\end{thebibliography}
\end{document}